%% file: iclr2024_conference.tex
\documentclass{article} 
\usepackage{iclr2024_conference,times}

\input{math_commands.tex}

\definecolor{vova}{rgb}{0.2,0.6,0.2}

\iclrfinalcopy

\usepackage{hyperref}
\usepackage{url}
\usepackage{graphicx}
\usepackage{cleveref}
\usepackage{subcaption}

\title{Towards Natural Language-Driven \\ Industrial Assembly Using Foundation \\ Models}

\author{Omkar Joglekar$^{*}$, Shir Kozlovsky$^{*}$, Vladimir Tchuiev$^{*}$, Zohar Feldman\thanks{All authors contributed equally to this work.} \\
Bosch Center for Artificial Intelligence\\
Haifa, Israel \\
\texttt{\{joo1tv, kos1tv, tcv1tv, fez1tv\}@bosch.com} \\
\And 
Tal Lancewicki$^{*}$\thanks{This work was done while the
author was an intern at the Bosch Center for Artificial Intelligence (BCAI).}\\
Tel Aviv University
\\
\texttt{lancewicki@mail.tau.ac.il}
\And
Dotan Di Castro \\
Bosch Center for Artificial Intelligence\\
Haifa, Israel \\
\texttt{did1tv@bosch.com}
}

\begin{document}

\maketitle
\begin{abstract}

Large Language Models (LLMs) and strong vision models have enabled rapid research and development in the field of Vision-Language-Action models that enable robotic control. The main objective of these methods is to develop a generalist policy that can control robots with various embodiments. However, in industrial robotic applications such as automated assembly and disassembly, some tasks, such as insertion, demand greater accuracy and involve intricate factors like contact engagement, friction handling, and refined motor skills. Implementing these skills using a generalist policy is challenging because these policies might integrate further sensory data, including force or torque measurements, for enhanced precision. In our method, we present a global control policy based on LLMs that can transfer the control policy to a finite set of skills that are specifically trained to perform high-precision tasks through dynamic context switching. The integration of LLMs into this framework underscores their significance in not only interpreting and processing language inputs but also in enriching the control mechanisms for diverse and intricate robotic operations.

\end{abstract}

\section{Introduction}

The advent of Large Language Models (LLMs) and strong vision models, triggered the development of strong Vision-Language Models (VLMs). These strong Foundation Models (FMs) are powered by the strength and versatility of the transformer model \citep{Vaswani2023AttentionNeed}. More recently, in the dynamic realm of industrial robotics, works based on these FMs, such as Octo, RT-X, PaLM-E \citep{GhoshOcto:Trajectories,Collaboration2023OpenModels, Driess2023PaLM-E:Model} showcases sophisticated technological prowess and heightened operational efficiency. This signifies a radical overhaul in robotics, where theoretical models are skillfully intertwined with practical robotic applications, thereby unlocking new, unprecedented capabilities. 

Specific methods such as Octo \citep{GhoshOcto:Trajectories}, RT-2 \citep{Brohan2023RT-2:Control}, PaLM-E \citep{Driess2023PaLM-E:Model} and ChatGPT for Robotics \citep{Vemprala2023ChatGPTAbilities} showcase ingenious methods to decode the complex learned tokens in LLMs such as T5 \citep{Raffel2019ExploringTransformer}, GPT-3 \citep{Brown2020LanguageLearners} and PaLM \citep{Chowdhery2022PaLM:Pathways}, to generate control actions for robots with minimal fine-tuning. This expansion from the initial fortes of LLMs in language processing to the complexities of robotic contexts illustrates their immense potential. The versatility and robustness of LLMs are crucial in reshaping robot manipulation, navigation, and interaction, presenting groundbreaking solutions to tasks previously deemed unfeasible. The unique challenges these models encounter, such as data scarcity, task variability, and the demand for real-time responsiveness, are extensively explored by \cite{Wang2024LargePerspectives} in their survey. This survey offers a thorough insight into the current landscape and future prospects of foundation models and LLMs in robotics.

\begin{figure}[th]
    \centering
    \includegraphics[width=\textwidth]{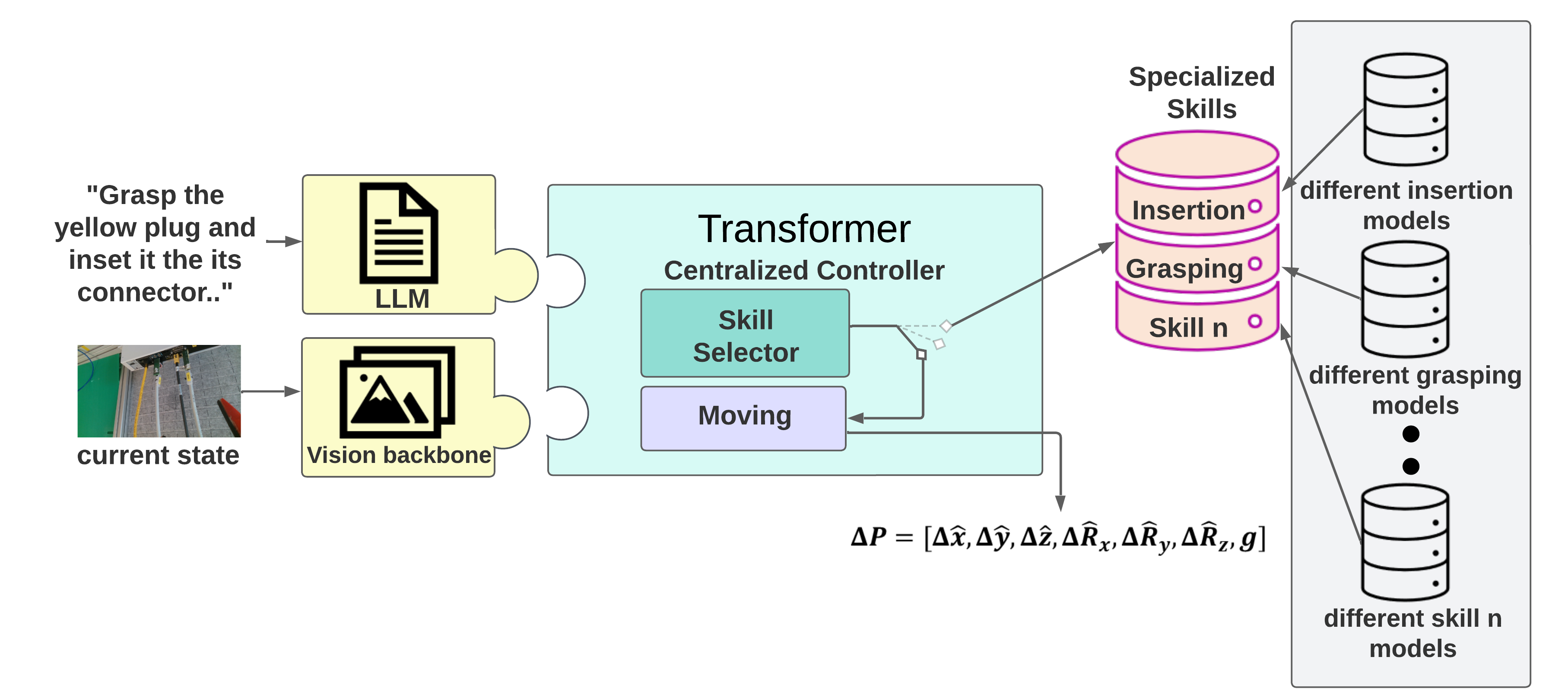}
    \caption{We introduced a user-friendly industrial assembly model based on natural language prompt and LLM, which is easily adaptable to new environments and modular in adding or adjusting components to meet specific needs. The modular nature is evident in the system's inputs and outputs. Regarding input flexibility, the model is designed to support various types of specialized skills' NN models}
    \label{fig:general-method}
\end{figure}

The implications of these advancements are especially significant in industrial assembly, where robots are expected to perform complex, high-precision tasks. Foundation models, with their enhanced decision-making, planning, and execution abilities, are pivotal for the intricate demands of industrial assembly. These models herald a shift towards smarter, more adaptive, and more efficient industrial processes, enabling robots to tackle complex and unstructured tasks that surpass the limitations of conventional automated systems. Our proposed method presents an approach that processes natural language-based goals from an assembly pipeline, such as "Carefully grasp the yellow plug and insert it into the appropriate socket." to output control instructions based on image observations.
In principle, our solution comprises a centralized goal-conditioned model that initiates readily available independent specialized skills by handing over the control to specialized models based on context switching. The centralized module is responsible for inferring the skill to be executed and moving the robot to an initial pose, enabling the specialized control model (skill model) to perform precise manipulation.
Specifically, the model outputs two control signals, namely:
\begin{enumerate}
    \item 7 Degree of Freedom (7DoF) pose for the default "move" action.
    \item Skill-based context class.
\end{enumerate}
The summary of our presented method can be seen in Fig~\ref{fig:general-method}. The "skills" mentioned above are small pre-trained networks specializing in fine-grained control tasks such as insertion \citep{Fu2022SafeInsertion, Spector2022InsertionNetInput}. We use the fine-grained skill "grasping for insertion", as an example of specialized skill, to test proper context switching. 
Our method is modular in terms of the text and image encoders and the type of policy used (we use a simple Multi-Layer Perceptron). In addition, our model is modular regarding specialized skills by simply fine-tuning the context classifier and calling the relevant policy to execute.

\section{Related Work}

\subsection{Vision-Language-Action modeling}
Multi-modal data processing has emerged as a new trend of research in the deep learning community. This is enabled by the versatility of the transformer architecture \citep{Vaswani2023AttentionNeed} to process tokens from various data sources. The versatility of transformers is highlighted in works such as ViLT \citep{Kim2021ViLT:Supervision} and Flamingo \citep{Alayrac2022Flamingo:Learning}, where tokens from images and text are successfully processed by a large transformer model for various downstream applications. Until recently, Vision-Language Models (VLMs) were studied rigorously for various applications. Notably, works such as VisualBERT, ViLBERT, MetaLM \citep{Li2019VisualBERT:Language, Lu2019ViLBERT:Tasks,Hao2022LanguageInterfaces} and others, as outlined by \cite{Gan2022Vision-LanguageTrends}, are enabled by accelerated research in large language models (LLMs) such as \cite{DevlinBERT:Understanding, Brown2020LanguageLearners, OpenAI2023GPT-4Report, Raffel2019ExploringTransformer} and vision models like \cite{Dosovitskiy2020AnScale}. VLMs are applied in multiple domains like visual question answering \cite{Zhou2019UnifiedVQA, Zellers2021MERLOT:Models}, captioning \cite{HuScalingCaptioning}, optical character recognition \cite{Li2021TrOCR:Models}, and object detection \cite{Chen2021Pix2seq:Detection}. 

More recently, embodied VLMs, also known as Vision-Language-Action (VLA) models, have gained popularity. These models can perform complex tasks using visual observations conditioned on text prompts. Works such as those introduced by \cite{Ahn2022DoAffordances, Driess2023PaLM-E:Model, Ha2023ScalingAcquisition, Hu2023LookPlanning} augment VLMs by adding support for robotic actions. In parallel, developing expansive models for precise low-level control is an actively researched field, with significant contributions by \cite{Brohan2022RT-1:Scale, Brohan2023RT-2:Control}. These models face challenges related to scalability and data requirements, as discussed by\cite{Collaboration2023OpenModels} and \cite{Kalashnikov2021MT-Opt:Scale}. These challenges highlight the need for a training framework that balances actionable outputs with scalability. Recent studies, such as those by \cite{Reed2022AAgent}, \cite{Yang2023Polybot:Variability}, and \cite{Kumar2023RoboHive:Learning}, have focused on models that map robot observations directly to actions, exhibiting impressive zero-shot or few-shot generalization capabilities across various domains and robotic systems. Notably, \cite{Bousmalis2023RoboCat:Manipulation} have made strides in managing different robot embodiments for goal-conditioned tasks. The RT-X model by \cite{Collaboration2023OpenModels} demonstrates proficiency in language-conditioned manipulation tasks across multiple robot embodiments, signaling a move towards more adaptable robotic systems. Very recently introduced by \cite{GhoshOcto:Trajectories}, Octo enables VLAs for multiple action spaces by leveraging the versatility of the transformer architecture. This modularity allows the adaptation of robotic FMs to different robot types by swapping the encoders and decoders depending on the embodiment of the robot.

Our work specifically addresses the challenge of converting language and vision inputs into robotic actions, pushing the boundaries of Transformers in robotic control. This extends beyond text-based conditioning, as explored in studies by \cite{Gupta2022MetaMorph:Transformers}, \cite{Jang2022BC-Z:Learning}, and \cite{Jiang2022VIMA:Prompts}, demonstrating the models' effectiveness across various robotic designs and task specifications.
Distinguishing our research, we have developed a centralized foundational model that combines multi-modality with pre-trained large models for specific sub-skills and tasks and offers flexibility in incorporating new skills or tasks. This model effectively connects to language commands and maintains robustness through feature modulation, marking a significant advancement in using Foundation Models in robotics.

\subsection{LLM In Robotics}

With recent success LLMs, most notably the GPT family by OpenAI \citep{Brown2020LanguageLearners, OpenAI2023GPT-4Report}, and the open-source Llama 2 model by Meta \citep{Touvron2023LlamaModels}, researchers aim to leverage the leap in reasoning capability for robotics applications. Examples include LM-Nav \citep{ShahLM-Nav:Action}, which tackles the problem of robotic navigation from language commands by utilizing CLIP for image and text interface, GPT-3 for text-based reasoning, and ViNG \citep{Shah2021Ving:Goals}, a model for visual navigation. Similarly, VLMaps \citep{Huang2022VisualNavigation} utilizes LLMs to translate natural-language commands to a sequence of navigation goals for a robot. \cite{Liang2023CodeControl} utilized an LLM fine-tuned on code data to create policies for robotic manipulation tasks via Python scripts that call task-based APIs. \cite{SongLLM-Planner:Models} proposed an LLM-based close-loop planning scheme for mobile robots for robust execution of multi-step tasks. ProgPrompt \citep{Singh2023ProgPrompt:Models} tackles in detail prompting for LLMs for planning and robotic tasks with a programmatic approach.

In combination with VLA models, LLMs are used for visual-based reasoning and planning.
VLA models such as RT-1 \citep{Brohan2022RT-1:Scale}), RT-2 \citep{Brohan2023RT-2:Control}, and PaLM-E \citep{Driess2023PaLM-E:Model} use tokens from pretrained LLMs as a major component in the VLA model and show impressive results. Other works, such as ChatGPT for Robotics \citep{Vemprala2023ChatGPTAbilities}, use GPT-3 \citep{Brown2020LanguageLearners} and clever prompting to generate actions for robots in simulated environments. Additionally, FMs, specifically LLMs, and VLMs, see active research for direct and actionable guidance, a topic that \cite{Huang2023VoxPoser:Models} have explored. Still, modeling real-world physical dynamic systems remains an open challenge. \cite{GhoshOcto:Trajectories} use T5-based \citep{Raffel2019ExploringTransformer} encoder to process language instructions. 

One major consideration in this design is the ability to update the LLM used in the VLA model with ease and minimal fine-tuning. All the aforementioned VLA models introduce various techniques to address this issue. Our method also allows effective swapping of the LLM, which ensures the longevity of the method in light of drastic LLM improvements in the future. In our method, we consider this as an important design decision to ensure effective future-proofing. 

\section{Methods}
\label{methods}

We introduce a centralized control model, employing the transformer architecture, that adeptly switches between specialized control skills from a predefined set, guided by natural language objectives and vision-based inputs.

This centralized controller fulfills two primary functions:
\begin{enumerate}
    \item Direct the robot to a specified location based on the text prompts
    \item Identify and predict the necessary specialized skill, such as grasping or insertion, based on the textual prompt and the robot's current state
\end{enumerate}
The first function, which we term the general "moving" skill, doesn't necessitate a highly precise 6 Degrees of Freedom (6DoF) pose estimation. The specialized tasks mentioned in the second function demand greater accuracy and involve intricate factors like contact engagement, friction handling, and refined motor skills. Additionally, they might integrate further sensory data, including force or torque measurements, for enhanced precision. Distinct from our core model, these specialized skills are developed independently, utilizing data specifically tailored to their requirements.

In this preliminary version of our work, we assume that these special skills work accurately, given that the robot meets certain constraints, such as being placed in an initial position that is in the proximity region of the manipulated object. The goal is be specified using a natural language prompt, for example, "Carefully grasp the yellow plug and insert it into the appropriate socket."

\begin{figure}[t]
    \centering
    \includegraphics[width=0.9\textwidth]{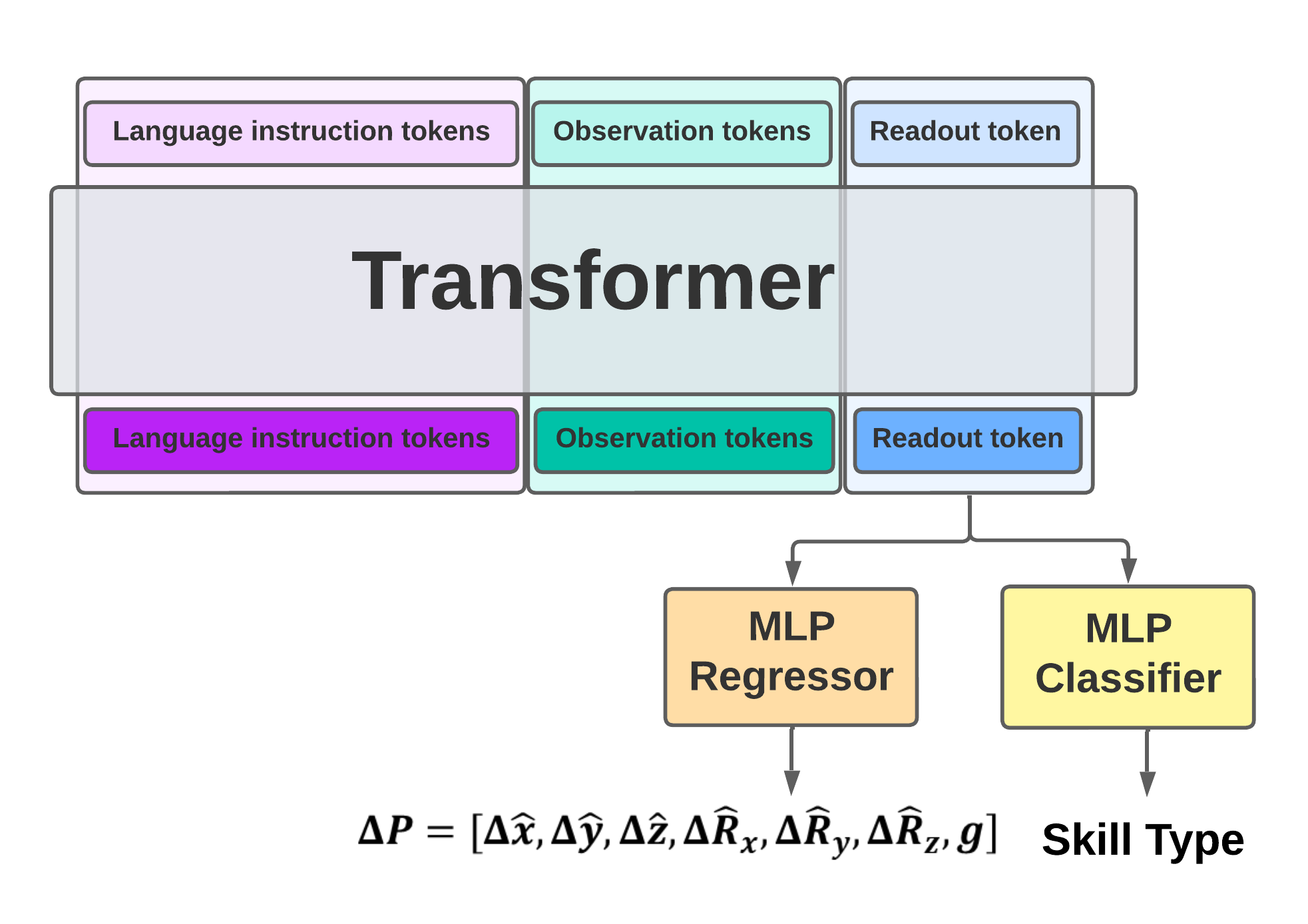}
    \caption{The centralized controller architecture: In each time step, Language tokens, Observation tokens, and the Readout token are concatenated together, where a mask is applied to ensure that the readout token attends all other tokens but is unattended by them. The MLP classifier then determines the appropriate skill to execute, and if the moving skill is selected, a small step is taken in accordance with the MLP Regressor's output.}
    \label{fig:transformer}
\end{figure}

As outlined in Fig.~\ref{fig:transformer}, our transformer model accepts language instruction tokens that are encoded by strong language models such as T5 \citep{Raffel2019ExploringTransformer}, BLIP \citep{Li2022BLIP:Generation} and CLIP \citep{Radford2021LearningSupervision}, that are pre-trained, frozen, and specialize in text encoding, and generate text instruction tokens. In addition, we use pre-trained vision encoders, such as ResNet-50\citep{He2015DeepRecognition} or ViT \citep{Dosovitskiy2020AnScale}, to generate vision tokens that embed information from the observations. For a more detailed discussion on the effect of each chosen encoder model, please refer to Section~\ref{ablations}. We pad the input with learnable "readout tokens", as described in Octo \citep{GhoshOcto:Trajectories}. In our implementation, the transformer implements a Markovian policy, wherein the action depends solely on the current observation and is independent of past observations. In alignment with our dual-purpose model, we bifurcate the action into two categories: the skill action, denoted as $a_s$, which pertains to the type of skill being executed, and the moving action, denoted as $a_m$, which relates to the movement skill. The problem can be formally defined as follows:

\begin{equation}
    \begin{aligned}
        a_s &= \pi_s(s), \\
        a_m &= \pi_m(s),
    \end{aligned}
\end{equation}

where $s$ is the state vector that encodes information about the current state ($image(t)$) and the general text prompt.

As described in Fig.~\ref{fig:transformer}, both policies largely share their weights and architecture but differ in their decoder models. Both policies mentioned are deterministic and based on the Multi-Layer Perceptron(MLP) architecture. The policy $\pi_s$ functions as a high-level controller, predicting the required skill by classifying predefined skills as follows:

\begin{enumerate}
    \setcounter{enumi}{-1}
    \item Terminate
    \item Moving (handled by the centralized controller)
    \item Skill 1 (specialized)
    \item Skill 2 (specialized)
    \item Skill 3 (specialized)
    \item[\vdots]
\end{enumerate}

"Terminate" indicates that the robot has reached its goal per the provided text prompt. When $a_s = \text{skill } n$, the control is handed over to the model specialized in skill $n$. When $a_s$ predicts the "moving" skill (denoted as "1"), the low-level controller's ($\pi_m$) action is executed. Additional specialized skills can be integrated by adding another context class and fine-tuning the classification head with data pertinent to the new skill.

The action space of $a_m$ is defined as a 7-dimensional vector, trained to predict a unit vector in the direction of the delta ground truth of the desired object or task using MSE loss. It is formulated as:
\begin{equation}
\Delta \mathbf{p} = [\Delta{x}, \Delta{y}, \Delta{z}, \Delta{R_x}, \Delta{R_y}, \Delta{R_z}, g]
\end{equation}

In this formulation, $\Delta{x}, \Delta{y}, \Delta{z}$ represent the translation components, while $\Delta{r_x}, \Delta{r_y}, \Delta{r_z}$ denote the orientation components represented in axis-angles, and $g$ corresponds to the opening of the gripper. This 7-dimensional vector is trained in a supervised manner using the Mean Squared Error (MSE) ($\mathbf{L}_{mse}$). 

We define \emph{active domain} as the region that enables the successful execution of a specialized skill. The boundary of this active domain is assumed to be an abstract threshold $\varepsilon$. This threshold varies for different skills and is not solely dependent on distance. For instance, when guiding a grasped plug to a socket for insertion, the context should revert to the "grasping for insertion" specialized skill if the plug's position becomes unfavorable for insertion. A classifier head is trained to estimate this abstract threshold and facilitate context switching accordingly. This multi-class classifier head is trained using Categorical Cross Entropy loss ($\mathbf{L}_{ce}$).

\section{Experimental setup}
\label{expt}
We evaluate the performance of our method using three methods:
 \begin{enumerate}
     \item The quality of the move action - to evaluate the performance of $\pi_m$.
     \item The accuracy of context switching - to evaluate how accurately the high-level controller transfers from one skill to another
     \item The overall system performance - This preliminary version of our work focuses on evaluating the performance of the full systems only for grasping tasks. It measures the system's ability to accurately identify and approach the right plug according to the text prompt, engage our independently trained grasping network, and grasp the plug correctly. We only present qualitative results for this experiment, and the presentation of this specialized skill will be included in future works.
 \end{enumerate} 
 The first method evaluates the quality of the moving action. Specifically, we measure whether the high-level control can bring the gripper within the active domain of the "grasping for insertion" skill. The second method evaluates whether the model accurately switches context when it is inside the active domain to perform "grasping for insertion". For this particular skill, the constraints can be summarized by Eq~\ref{eq:c1} and~\ref{eq:c2} 
 \begin{equation}
 \label{eq:c1}
     \|[\Delta{x}, \Delta{y}, \Delta{z}]\| \le 4 [cm]
 \end{equation}
 \begin{equation}
 \label{eq:c2}
     \|[\Delta{r_x}, \Delta{r_y}, \Delta{r_z}]\| \le 20 [deg]
 \end{equation}

\begin{figure}[ht]
\centering
\begin{subfigure}{.33\textwidth}
  \centering
  \includegraphics[width=.95\linewidth]{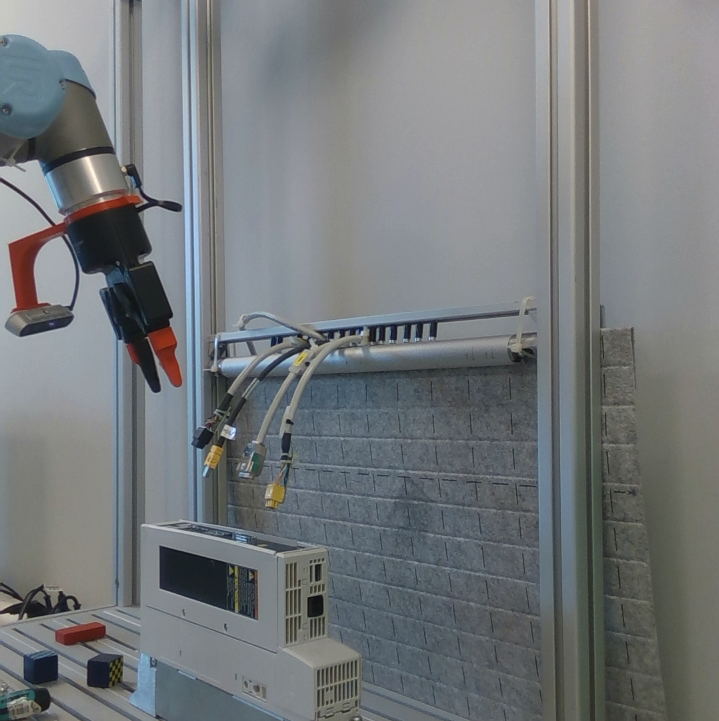}
  \caption{Baseline scene}
  \label{fig:sub1}
\end{subfigure}%
\begin{subfigure}{.33\textwidth}
  \centering
  \includegraphics[width=.95\linewidth]{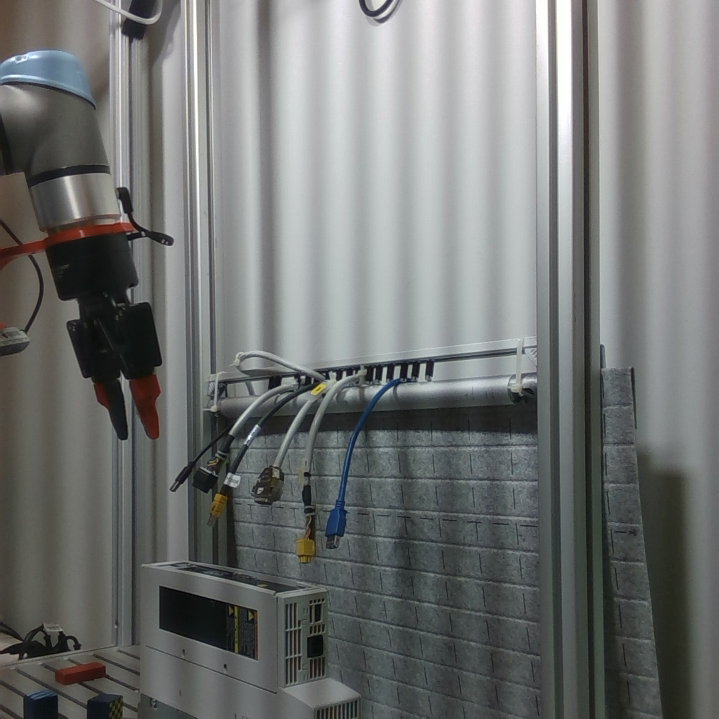}
  \caption{Plug Distractions}
  \label{fig:sub2}
\end{subfigure}%
\begin{subfigure}{.33\textwidth}
  \centering
  \includegraphics[width=.95\linewidth]{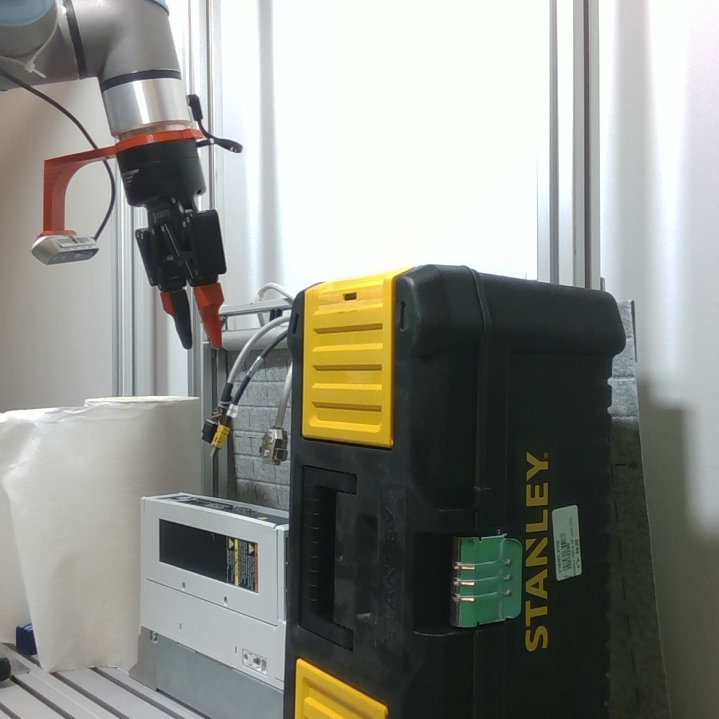}
  \caption{Object Distractions}
  \label{fig:sub3}
\end{subfigure}

\begin{subfigure}{.33\textwidth}
  \centering
  \includegraphics[width=.95\linewidth]{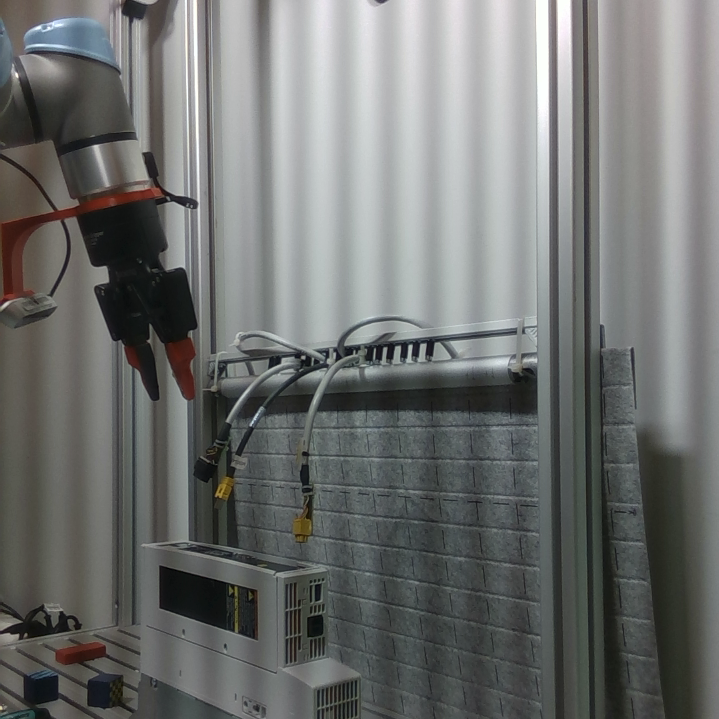}
  \caption{Missing Plug}
  \label{fig:sub4}
\end{subfigure}%
\begin{subfigure}{.33\textwidth}
  \centering
  \includegraphics[width=.95\linewidth]{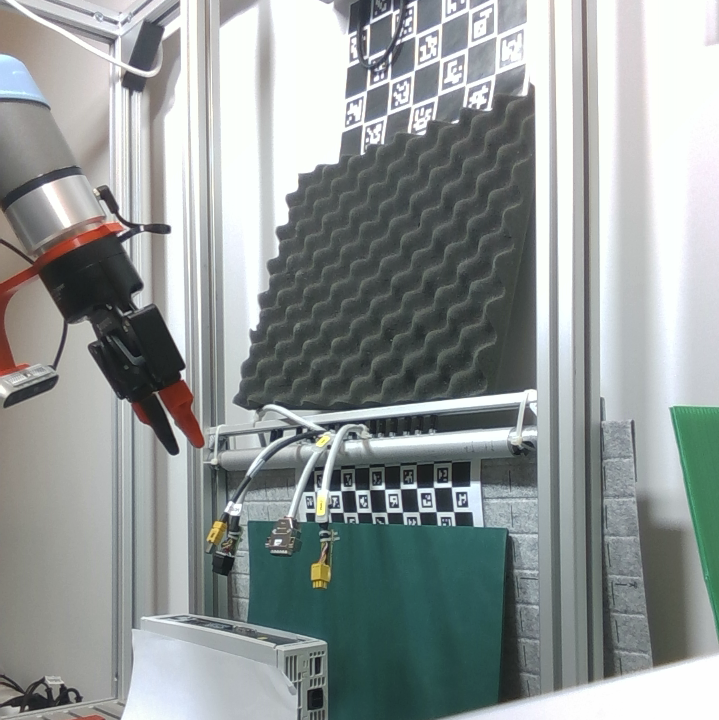}
  \caption{Unseen Background}
  \label{fig:sub5}
\end{subfigure}

\caption{Examples of perturbations introduced in the experimental setup to evaluate the robustness of our model. Each sub-figure illustrates a different type of perturbation.}
\label{fig:perturbs}
\end{figure}

To evaluate the robustness of our model, we introduce several perturbations to our experimental setup:
\begin{itemize}
    \item Plug Distractions: In addition to the four plugs initially trained on, we introduce two additional plugs of a different type into the scene to act as distractions.
    \item Object Distractions: We include large objects not present in the training scenes.
    \item Missing Plug: We remove one of the four plugs to examine if its absence influences the success rate for the remaining plugs.
    \item Unseen Background: We modify the background to an unseen variant.
\end{itemize}
Example scenarios for each perturbation can be seen in Fig.~\ref{fig:perturbs}

\section{Results}
\subsection{Quantitative Results}
Table~\ref{tab:success_rates} shows how many times the moving policy $\pi_m$ successfully brings the gripper into the active domain (as described in Section~\ref{expt}) for the skill. We observe that the model performs well on the plugs within the model's training set.
\begin{table}[h]
    \centering
    \begin{tabular}{|l|c|}
        \hline
        \textbf{Perturbation} & \textbf{Success Rate} \\ \hline
        Baseline (No Perturbation) & $33/40 (82.5\%)$ \\ \hline
        Plug Distractions & $9/19 (47\%)$ \\ \hline
        Object Distractions & $20/23 (87\%)$ \\ \hline
        Missing Plug & $26/36 (72\%)$ \\ \hline
        Unseen Background & $12/17 (70.6\%)$ \\ \hline
    \end{tabular}
    \caption{Success Rates of Central Controller Under Various Environment Perturbations}
    \label{tab:success_rates}
\end{table}

\begin{figure}[ht]
  \centering
  \begin{subfigure}{0.3\linewidth}
    \centering
    \includegraphics[width=\linewidth]{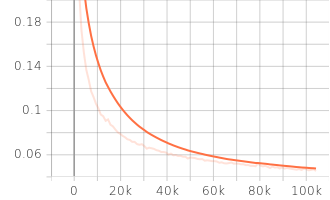}
    \caption{Train Loss}
    \label{fig:train-loss}
  \end{subfigure}
  \begin{subfigure}{0.3\linewidth}
    \centering
    \includegraphics[width=\linewidth]{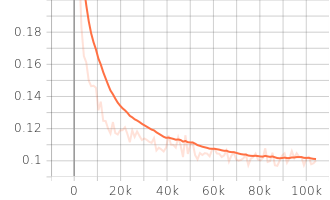}
    \caption{Validation Loss}
    \label{fig:val-loss}
  \end{subfigure}
  \\
  \begin{subfigure}{0.3\linewidth}
    \centering
    \includegraphics[width=\linewidth]{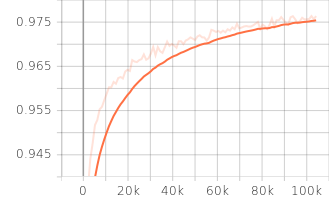}
    \caption{Context Switching accuracy (Train)}
    \label{fig:train-acc}
  \end{subfigure}
  \begin{subfigure}{0.3\linewidth}
    \centering
    \includegraphics[width=\linewidth]{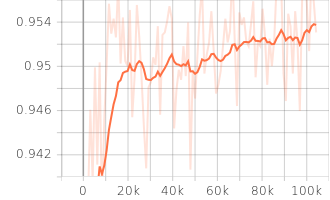}
    \caption{Context Switching accuracy (Validation)}
    \label{fig:val-acc}
  \end{subfigure}
  \caption{Convergence Graphs and Context Switching Accuracy}
\label{fig:subfigures}
\end{figure}

The context switching is modeled as a multi-class classification problem (Section~\ref{methods}). The Fig.~\ref{fig:train-acc} and Fig.~\ref{fig:val-acc} show the accuracy of this multi-class classification. We see that the final context switching accuracy is $\approx95\%$ in validation.

\subsection{Qualitative Results}
In the additional materials, we upload videos of the prediction results of the trained model. We present demo videos of reaching and the "grasping for insertion" skill. All videos can be found in the project website: \url{https://sites.google.com/view/roboticfm}. The detailed description of the "grasping for insertion" skill is outside the scope of this work and will be presented in future methods. 

\section{Ablation Study}
\label{ablations}

The major contribution of our work is a unified transformer model that fuses language and image data to generate control signals for robotic operation. We performed experiments to choose the most effective transformer architecture. Recently as a part of the Segment Anything Model (SAM) \cite{Kirillov2023SegmentAnything}, the authors introduced a new attention mechanism termed as "two-way attention", in their mask decoder. This attention mechanism differs from regular cross-attention mechanisms in that it updates the context and inputs in each layer. We refer curious readers to this paper for more information about this technique. Motivated by their success in performing complex segmentation tasks, we also test a transformer network that uses a stack of two-way attention blocks described in SAM, along with the standard transformer-decoder network that uses a stack of self-attention blocks (\cite{Vaswani2023AttentionNeed}).

While using the two-way attention layers, we input the image embeddings as the "dense tokens" and the readout and language tokens as "sparse tokens". The presence of two cross-attention layers in each such layer drastically increases the memory requirement and processing time. We set the depth of the transformers such that both mechanisms have a similar number of parameters ($\sim 20M$). We find that using a two-way transformer of the same size does not generalize well to points outside the dataset and drastically reduces success rates.

The inherent modularity presented in our model opens up a discussion about the choice of image and text encoders. Broadly speaking, we test two types of text encoders: 
\begin{enumerate}
    \item Encoders pre-trained on Vision-Language tasks such as CLIP or BLIP text encoders
    \item Encoders pre-trained Language generation such as the suite of T5 text encoders
\end{enumerate}
Both the T5-small and CLIP text encoders show promising performance. However, CLIP outperforms the T5-small model due to its experience with pairing images to text. The results presented in~\ref{tab:success_rates} are for the model that uses CLIP.
The size of the text encoder affects the stage in the pipeline where the data fusion begins. In other words, if the text encoder is very deep, then the image-language fusion begins much later on in the pipeline. This effect becomes prominent when swapping the CLIP text encoder for the BLIP one, which is much deeper. Specifically, the model's performance drops when we use the BLIP text encoder. All the text encoder tests were done while keeping the same image encoder (ResNet-18 pre-trained on ImageNet).

To determine the best image encoder, we fix the CLIP text encoder. We try using ResNet-50 pre-trained on ImageNet and ResNet-50 pre-trained as a part of CLIP. Unsurprisingly, the latter image encoder shows better performance, generalization, and robustness. Table~\ref{tab:success_rates} presents results using this type of image encoder.

\section{Conclusions and Future Work}

This work presents a Robotics Foundation Model that uses text-based task specification and image-based observations to generate control actions for robotic assembly tasks. The highlighting features of our model include modularity, ease of use, and the ability to add any finite number of specialized skills, such as insertion, to the \emph{skill library} that the centralized model can control. We present a comprehensive architecture, training protocol, and qualitative \& quantitative results of the described method.

We wish to emphasize that this project is still in the research phase. In future works, we aim to introduce a bigger database of compatible skills and support for task specification using goal images. Success rate improvement is also an active research topic, and implementing a full pipeline for more complex task specifications to deal with collisions and occlusions is also under exploration. We also plan to add datasets from various robotic setups and assembly lines to improve the generalization ability for additional assembly tasks.

\bibliography{iclr2024_conference}
\bibliographystyle{iclr2024_conference}


\end{document}

%% file: math_commands.tex
\usepackage{amsmath,amsfonts,bm}

\def\eqref#1{equation~\ref{#1}}

\def\1{\bm{1}}

\DeclareMathAlphabet{\mathsfit}{\encodingdefault}{\sfdefault}{m}{sl}
\SetMathAlphabet{\mathsfit}{bold}{\encodingdefault}{\sfdefault}{bx}{n}

